\documentclass[letterpaper, 10 pt, conference]{ieeeconf}  
\IEEEoverridecommandlockouts                              

\usepackage[utf8]{inputenc} 
\usepackage[T1]{fontenc} 
\usepackage{cite} 
\usepackage{graphics} 
\usepackage{times} 
\usepackage{amsmath} 
\usepackage{amssymb} 
\usepackage{amsfonts}
\usepackage{mathtools} 
\usepackage{graphicx}
\usepackage{textcomp}
\usepackage{tikz}
\usepackage{siunitx} 
\usepackage{wrapfig}
\usepackage{tabularx}
\usepackage{booktabs} 
\usepackage{makecell}
\usepackage{multirow} 
\usepackage{hhline}
\usepackage{tikz}
\usetikzlibrary{backgrounds, fit, shapes.geometric, positioning}
\usepackage{url}
\usepackage{xcolor}
\usepackage{soul}
\usepackage{rotating}
\usepackage[hidelinks,bookmarksnumbered,bookmarksdepth=1]{hyperref} 

\usepackage[ruled, linesnumbered, vlined, noend]{algorithm2e}

\SetCommentSty{mycommfont}
\usepackage[noend]{algpseudocode}

\usepackage{balance} 
\def\secref#1{Sec.~\ref{#1}}
\def\figref#1{Fig.~\ref{#1}}

\def\tabref#1{Tab.~\ref{#1}}

\def\eqref#1{Eq.~(\ref{#1})}

\algnewcommand{\algorithmicgoto}{\textbf{go to}}
\algnewcommand{\Goto}[1]{\algorithmicgoto~line~\ref{#1}}

\newcommand\etal{~\emph{et al.}}

\newcolumntype{Y}{>{\centering\arraybackslash}X}
\newcommand{\ccdot}{\!\cdot\!} 

\title{\LARGE \bf Compact Multi-Object Placement\\Using Adjacency-Aware Reinforcement Learning}
\author{Benedikt Kreis$^{1,2,3}$ \and Nils Dengler$^{1,2,3}$ \and Jorge de Heuvel$^{1,2,3}$ \and Rohit Menon$^{1,2}$ \and Hamsa Perur$^{1}$ \and Maren Bennewitz$^{1,2,3}$
  \thanks{
  	\hspace{-1.05em}$^{1}$Humanoid Robots Lab, University of Bonn, Germany.\newline
  	$^{2}$Center for Robotics, University of Bonn, Germany.\newline
  	$^{3}$Lamarr Institute for ML and AI, Bonn, Germany.\newline
         This work has been partially funded by the EC, grant No. 964854 RePAIR H2020-FETOPEN-2018-2020 and by the BMBF within the Robotics Institute Germany, grant No. 16ME0999.\newline
	The corresponding author is Benedikt Kreis: \href{mailto:kreis@cs.uni-bonn.de}{kreis@cs.uni-bonn.de}}%
}

\begin{document}
\maketitle
\thispagestyle{empty} 
\pagestyle{empty}

\begin{abstract} 
Close and precise placement of irregularly shaped objects requires a skilled robotic system.
The manipulation of objects that have sensitive top surfaces and a fixed set of neighbors is particularly challenging.
To avoid damaging the surface, the robot has to grasp them from the side, and during placement, it has to maintain the spatial relations with adjacent objects, while considering the physical gripper extent.
In this work, we propose a framework to learn an agent based on reinforcement learning that generates end-effector motions for placing objects as closely as possible to one another.
During the placement, our agent considers the spatial constraints with neighbors defined in a given layout of the objects while avoiding collisions.
Our approach learns to place compact object assemblies without the need for predefined spacing between objects, as required by traditional methods.
We thoroughly evaluated our approach using a two-finger gripper mounted on a robotic arm with six degrees of freedom.
The results demonstrate that our agent significantly outperforms two baseline approaches in object assembly compactness, thereby reducing the space required to position the objects while adhering to specified spatial constraints.
\end{abstract} 

\section{Introduction}
\label{sec:intro}
%
%
%
Object manipulation, particularly picking and placing, is a highly researched field in robotics~\cite{lobbezooReinforcementLearningPick2021a, kleebergerSurveyLearningBasedRobotic2020, leePickPlaceTacklingRobotic2022,zhangGraspStackingDeep2020a}.
The different applications lead to a variety of challenges such as limited workspace, tight tolerances affecting error margins, object quantity, and shape diversity.
Thus, one of the main problems is the precise positioning of objects at target poses \cite{berscheidSelfSupervisedLearningPrecise2020a}, which is especially difficult when dealing with restrictions of touching the top surface, e.g., with a suction gripper, due to sensitive or irregular surfaces, necessitating top-down, side grasping.
This scenario arises in sectors such as electronics production, automotive manufacturing, or also archaeology in the reconstruction of historical artifacts.

\begin{figure}[t]
	\centering
	\includegraphics[width=0.99\linewidth]{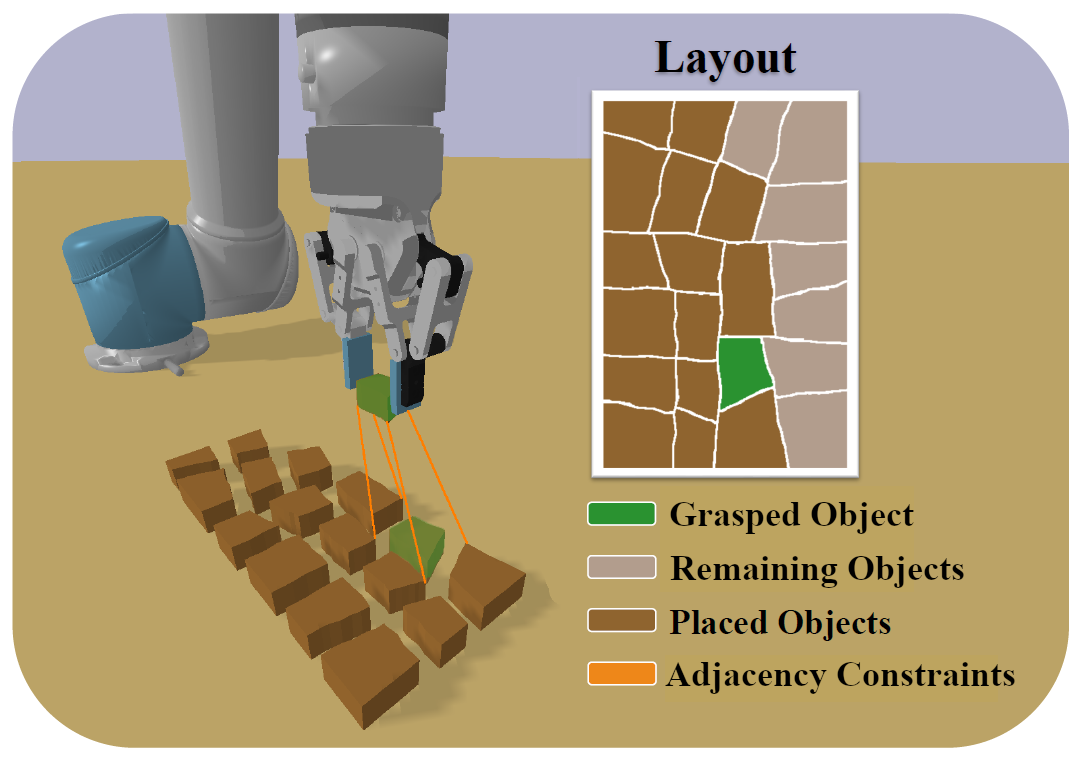}
	\caption{The goal of our approach is to place all objects as compactly as possible to one another while considering the objects' adjacency constraints.
	}
	\label{fig:motivation}
\end{figure}

%
%
The aforementioned challenges are partially tackled by dense packing methods \cite{wangDenseRoboticPacking2022}, \cite{renAutonomousOrePacking2024}.
However, current work in this field has its limitations and does not consider, e.g., contactless, adjacent relations between objects, or the need for sufficient space between the objects for the gripper's operation due to the top-down, side-grasping restriction.

%
%
In this paper, we propose a reinforcement learning (RL) approach to enable robotic manipulators to place objects closely together according to a layout that defines adjacency constraints between the objects.
Our approach learns to maximize the resulting assembly compactness while ensuring a collision-free placement of all objects.
From an initial grasping position, our agent tries to place an arbitrary shaped object as close as possible to its desired position according to a given layout while considering adjacency constraints to previously placed objects.
By design, our RL agent is able to handle sub-optimal grasps resulting from arbitrary yaw orientations of the object during the picking process.

\figref{fig:motivation} shows our agent in the process of placing an exemplary assembly by controlling the robotic manipulator, in this case, a robotic arm with six degrees of freedom (DoF) and a two-finger gripper.
As can be seen, the robot places the grasped object into the assembly, while avoiding collisions and minimizing the distances to the previously placed neighboring objects.
As our experimental results show, our approach varies the required space between the objects for collision-free gripper operation depending on their shape, consequently reducing the required assembly area of all placed objects by more than \SI{70}{\percent} compared to two baseline approaches.

To summarize, the main contributions of our work are: A framework for compact multi-object placement of irregularly shaped objects, that maintains spatial constraints with adjacent objects while avoiding collisions, by learning end-effector (EE) motions using RL. Moreover, we conducted an ablation study, as well as qualitative and quantitative evaluations, comparing our approach to two baselines, which demonstrate its superior performance in terms of assembly compactness.

\section{Related Work}
\label{sec:related}
Research in pick-and-place (P\&P) manipulation involves a broad spectrum ranging from the entire P\&P pipeline to specific tasks such as object grasping and stacking \cite{liuBriefReviewRecent2022, parkKnowledgeBasedReinforcementLearning2023a, orsulaLearningGraspMoon2022, ceolaGraspPoseAll2023}.
The exploration of deep learning techniques, such as reinforcement learning, brought significant advances to this domain \cite{lobbezooReinforcementLearningPick2021a, kleebergerSurveyLearningBasedRobotic2020}.
Yet, only limited literature considers the distinct task of positioning objects side by side while optimizing object placement under consideration of given adjacency constraints and avoiding collisions.

In the field of self-supervised learning for precise object placement, Berscheid\etal~\cite{berscheidSelfSupervisedLearningPrecise2020a} presented a method that combines learning primitives with one-shot imitation learning to align objects according to a visual input.
While they use an explicit visual goal, our method does not need an explicit target pose for object placement and instead adaptively finds a desired configuration that maintains the spatial relation with adjacent objects while avoiding collisions.

Lin\etal~\cite{linEfficientInterpretableRobot2022} use a graph neural network (GNN) policy trained using imitation learning to learn spatial relations between objects to perform long range manipulation tasks.
However, their method treats the manipulation problem as a classification of what actions to take and they do not concentrate on the low level motion planning and placement.

In a related work integrating spatial information with GNNs using a relational RL architecture called ReNN, Li\etal~\cite{liPracticalMultiObjectManipulation2020} trained an RL agent with sparse rewards using curriculum learning for stacking cuboids.
In contrast to their approach, our approach does not require curriculum learning to converge.

Zhang\etal~\cite{zhangGraspStackingDeep2020a} demonstrated the effectiveness of deep Q-learning for stacking cuboids side by side, exploiting the uniformity of the cuboids for simplified grasping and placing. As they focus on stacking a wall-like structure, they can avoid collisions by opening the gripper orthogonal to the structure.
In contrast, our approach has to learn to avoid gripper collisions as the objects are arranged horizontally next to one another.

In Pick2Place~\cite{hePick2PlaceTaskAware6DoF2023a}, He\etal~estimate task-aware grasps using object-centric affordances for shelf placements and object insertion tasks. 
However, they do not consider relations between objects when placing them on the shelf, treating all objects as interchangeable as long as they fit.

Dense packing and the peg-in-a-hole task represent benchmarks for assessing the precision and efficiency of P\&P manipulations.
Studies in dense packing often simplify the complexity of the task by focusing on heuristic and RL-based strategies for fitting objects within limited spaces, without fully addressing diverse object shapes and gripper collisions \cite{wangDenseRoboticPacking2022, renAutonomousOrePacking2024}.
Furthermore, in dense packing, there is no specific order for sequential object placements, nor do the objects need to fulfill any spatial constraints with neighboring objects.
In contrast, our work focuses on compact multi-object placement in a specified assembly sequence while at the same time maintaining the desired relative poses between adjacent objects.
%
Even though the peg-in-a-hole task \cite{kimActiveExtrinsicContact2022} is challenging due to the need for high object placement accuracy for insertion, it does not address the challenges of adaptive, multi-object placement while avoiding collisions between objects. 

Our method improves on existing methodologies by learning to compactly place objects while avoiding collisions.
To the best of our knowledge, no prior work has focused on placing objects in a compact assembly without specifying explicit target poses while maintaining the spatial constraints between adjacent objects.

\section{Problem Definition}
\label{sec:assumptions}
We consider the following problem: In a plane environment, a robotic arm has to compactly place irregularly shaped objects according to a given layout specifying spatial constraints between adjacent objects.
We assume the usage of a two-finger end-effector, moving in the three translational axes and one rotational (yaw) axis \((x, y, z, \theta)\).
Since we control the EE, the robotic arm can be of any degree of freedom.
Our RL agent has to determine the best incremental movement \((\Delta x, \Delta y, \Delta z, \Delta \theta)\) of the EE at each time step to efficiently move the current object to the most suitable pose according to the constraints defined in the assembly layout.
Thereby, the closer the agent places the objects next to each other the better, while avoiding collisions.
Note that due to the necessary top-down, side grasping, the agent has to leave space to allow for a collision-free opening of the gripper.
We assume the placement sequence is analytically derivable from the given layout.
Furthermore, we assume that all object poses are reliably perceptible and that CAD models of the robot and all objects are available.

In the following, we distinguish between three types of objects: The \textit{Grasped Object}, which the robot has currently grasped for placing; the \textit{Placed Objects}, which it has already placed at their final positions in the workspace; and the \textit{Remaining Objects}, which it has not placed yet.

\section{Our Approach}
\label{sec:main}

Since the optimal spacing between objects for tight and collision-free placement is unknown, we propose an RL system that learns to place each object as close as possible to its neighbors according to the layout, while avoiding collisions with the objects and the gripper.

\begin{figure}[t] 	\centering 	\includegraphics[width=0.99\linewidth]{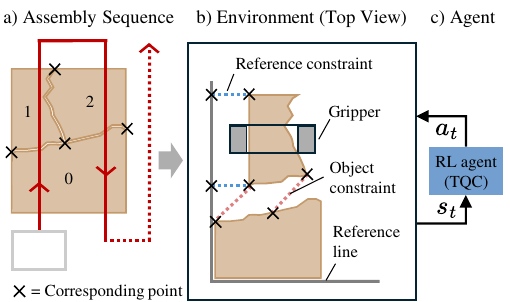}
\caption{
        Overview of our approach:
		a)~To determine the assembly sequence, we slide a virtual window with a fixed size in a snake-like pattern (red line) over the layout.
		Furthermore, we determine the corresponding points by searching for the two most distant points between all shared object edges in the layout.
		b)~The exemplary environment depicts two objects from a top-down view.
		The bottom one, the robot has already placed along the reference lines and it has grasped the next one in the assembly sequence, which is now between the gripper fingers.
		To maneuver the grasped object close to its neighbors, the RL agent uses the corresponding points to calculate the object and the reference constraints.
		c)~At each time step \(t\), the RL agent observes the state of the environment \(s_t\) and sends action commands \(a_t\) to the EE.
	} 
	\label{fig:architecture}
\end{figure}

As we focus on the placing action, for grasping, we use an automated routine that selects random, top-down grasp poses.
This way, our agent learns to place objects as best as possible, even with sub-optimal grasp poses.
\figref{fig:architecture} shows an overview of our approach.
Following the assembly sequence extracted from the layout with a sliding window approach (see \figref{fig:architecture}\,a and \secref{sec:generator} for details), our agent tries to place an arbitrarily shaped object closely to its neighbors that have been previously placed.
In the following sections, we introduce our approach in more detail.

\subsection{Object Neighbor Concepts}
\label{subsec:constraints}
%
To support the learning of our system, we introduce two concepts, used in the reward function and the observation space, that guide the agent to place objects close to each other.
As our exemplary assembly application is the reconstruction of broken fresco fragments (see \secref{sec:generator}), we took inspiration from techniques used by archaeologists, such as markings and construction lines \cite{enayatiSemanticMotifSegmentation2023}.
Similar to markings between corresponding fragments, we calculate the corresponding points \(c_i \in N_c\) on adjacent object polygons in the layout to constrain the placement of neighboring objects.
Each \(c_i\) corresponds to the outer-most points of shared edges between neighboring objects (see \figref{fig:architecture}\,a).
Furthermore, to align the objects in the workspace, we define two reference lines (\(l_x,l_y\)) located at the assembly's \(x\) and \(y\) axis, which serve as virtual construction lines (see \figref{fig:architecture}\,b).

\subsection{RL Task Description}
\label{subsec:episode}
During an episode, the goal of the agent is to place an object in the workspace, as shown in \figref{fig:motivation}.
At the beginning of each training episode, a routine initializes the robot to a start configuration with the EE \SI{5}{cm} above the workspace grasping a randomly selected object $o_i$ out of the layout.
Given the assembly sequence, we spawn all preceding objects $o_1, \dots, o_{i-1}$ in the workspace to simulate a successful placement of them.
Both, the grasp yaw angle of $o_i$ (\(\theta_{\text{grasp}}\)) and the EE's yaw angle \((\theta_{\text{EE}}\)) are randomly chosen according to (\(\theta_{\text{grasp}}, \theta_{\text{EE}} \in [-\SI{90}{\degree},+\SI{90}{\degree}]\)).
Afterwards, the agent has to place the object as close as possible to the ones that are already in the workspace.
An episode terminates unsuccessfully when the maximum number of steps \(N_\text{ep}\) is reached, or if a collision occurs.
When an episode terminates, we reset the environment and the agent starts again to place the next object.
Note that during training, the agent places single, randomly sampled objects.
During inference, we extend the task to place all objects sequentially according to the given assembly sequence.

\subsection{RL Architecture}
\label{subsec:architecture}
RL optimizes transitions from one state to the next \(s_t \rightarrow s_{t+1}\) following the Markov Decision Process.
The reward \(r_t = r(s_t, a_t)\) describes the incentive for the RL agent to take an action \(a_t = \pi_{\phi}(s_t)\) at time step \(t\) with respect to a policy \(\pi_{\phi}\).
Pairs of tuples \(\left(s_t, a_t, r_t, s_{t+1}\right)\) summarize states and actions.
The objective is to maximize the cumulative return \(R = \sum^{T}_{i=t} \gamma^{(i-t)}r_t\) of the \(\gamma\)-discounted rewards.

\subsubsection*{\bfseries{RL Algorithm}} 
\label{subsubsec:rl_algo}
In this work, we use the off-policy algorithm Truncated Quantile Critics (TQC) \cite{kuznetsovControllingOverestimationBias2020} with two critics and \(25\) quantiles.
To ensure sufficient exploration, we add Gaussian noise from a process \(\mathcal{N}\) with a standard deviation \(\sigma_{\epsilon_\pi}\) to the actions, so that \(a_t = \pi_{\phi}(s_t) + \mathcal{N}(0, \sigma_{\epsilon_\pi})\).

\subsubsection*{\bfseries{Action Space}} 
\label{subsubsec:action_space}
At each time step, our agent computes the next incremental action \((\Delta x,\Delta y,\Delta z,\Delta\theta)\) for the robot's EE. Each increment value has to be on the interval \([-\SI{1}{cm},+\SI{1}{cm}]\) for translations and on the interval \([-\SI{10}{\degree},+\SI{10}{\degree}]\) for the orientation.
Additionally, the agent decides when to open the gripper.
We normalize all actions to values on the interval \([-1,+1]\).

\subsubsection*{\bfseries{Observation Space}} 
\label{subsubsec:observation_space}
\begin{table}[b]
	\centering
	\caption{Observation space}
	\label{tab:observations}
	\begin{tabularx}{\linewidth}{l*{3}{Y}}
		\toprule[\lightrulewidth]
		\multicolumn{1}{c}{Observation} & \multicolumn{1}{c}{Notation} & \multicolumn{1}{c}{Size} \\
		\midrule[\lightrulewidth]
		Object-object point distances & \((x,y,z)_\text{o}\)  & 36 \\
		Reference-object point distances & \((x,y,z)_\text{l}\) & 12 \\
		EE pose & \((x,y,z,\theta)_\text{EE}\) & 4 \\
            Minimum gripper finger distance & \((x,y,z)_\text{gf}\) & 3 \\
            Minimum grasped object distance & \((x,y,z)_\text{g}\) & 3 \\
		Grasped object yaw angle & \(\theta_\text{g}\) & 1 \\
  		Grasped object overlap Boolean & \(\Omega_\text{g}\) & 1 \\
		\bottomrule[\lightrulewidth]
		& Sum & 60 \\
	\end{tabularx}
\end{table}

\tabref{tab:observations} shows the complete observation space.
The first two listed observations represent the three-dimensional distances between corresponding points of the grasped object and the reference lines as well as between the grasped object and the placed objects in its vicinity.
Note that the maximum number of expected object neighbors (\(max(N_\text{v})\)\,\(=\)\,\(6\)) defines the partial vector size for corresponding object-object point distances (\(N_\text{c}\)).
Since a grasped object cannot have more than two reference lines as its neighbors (first object), this amount defines the partial vector size for corresponding reference-object point distances (\(N_\text{l})\).
The EE pose is essential for the agent to know, as it reflects the result of each incremental movement it takes.
Furthermore, the agent has to be able to reason about collisions.
Therefore, it receives the three-dimensional minimum distances between the placed objects and the EE's fingers as well as the grasped object.
The grasped object orientation and the overlap Boolean aid the agent to find collision-free placements.
We calculate the overlap by checking if the two-dimensional shape representation of the grasped object intersects with any placed object in the vicinity. 
To mask unavailable or irrelevant observations, we pad them with~\(0\), e.g., for the first object in the assembly when no placed objects are present.
We normalize all observations to values on the interval~\([0,+1]\).

\subsection{Reward}
\label{subsec:reward}
The reward design takes three main objectives into account.
First, to efficiently move the grasped object to its neighbors in the workspace.
Second, to improve the compactness of the assembly.
Third, the object placement must be achievable without collisions.
The reward \(r\) consists of three components:
\begin{equation}
\label{eq:reward}
r = r_\text{m} + r_\text{r} + r_\text{col}
\text{.}
\end{equation}

The first two components incentivize the agent to \underline{m}ove the grasped object towards its neighbors while considering constraints via \(r_\text{m}\) and \underline{r}elease it close to its neighbors via \(r_\text{r}\).
The last component \(r_\text{col}\) penalizes the agent for \underline{col}lisions throughout the episode.
In the following paragraphs, we introduce all reward components in detail.
All \(\alpha\) and \(\beta\) are scaling factors.
While the former defines the magnitude, the latter shapes the slope of the corresponding partial reward component. 
We normalize all distances and angles used in the reward calculations to values on the interval \([0,+1]\).

\subsubsection*{\textbf{Constrained Object Movement}}
The agent has to move the given object towards its placement position in the environment.
During this movement, time-inefficient solutions result in a penalty for the agent according to the reward \(r_{m}\), which incentivizes it to move the grasped object as fast as possible towards its preceding neighbor objects or an adjacent reference line.
Furthermore, \(r_{m}\) ensures the correct object orientation \(\theta_\text{o}^\text{place}\) according to the layout \(\theta_\text{o}^\text{layout}\) with \( \Delta\theta_\text{o}\)\,\(=\)\,\(| \theta_\text{o}^\text{layout}\)\,-\,\(\theta_\text{o}^\text{place} |\).
We define \(r_{m}\) as:
\begin{equation}
\label{eq:reward_m}
r_\text{m} =
- \alpha_\text{n} \cdot
(
\underbrace{d_\text{o} + d_\text{l}}_{\text{constraints}} +
\underbrace{\Delta\theta_\text{o}}_{\text{yaw}}
)
\text{,}
\end{equation}

where \(d_\text{o}\) is the mean Euclidean distance of corresponding object-object pairs between the grasped object (\(i = 1\)) and all neighbor objects (\(i \in \{2,...,N_\text{v}\}\)), with \(N_\text{v}\) being the amount of neighbor objects in the vicinity.
Respectively, \(d_\text{l}\) is the mean Euclidean distance of corresponding reference-object pairs between all reference lines (\(l_\text{x},l_\text{y}\)) and the grasped object.
We calculate \(d_\text{o}\) with: 
\begin{equation}
\label{eq:object_dist}
d_\text{o} = 
\frac{
	\sum_{i=2}^{N_\text{v}} 
	\frac{	
		\sum_{k=1}^{N_\text{c}}
		|| \vec{p}_{1,k} - \vec{p}_{i,k} ||
	}{N_\text{c}}
}
{
	N_\text{v}
}
\text{,}
\end{equation}

where the vector \(\vec{p} = \big(\begin{smallmatrix} x\\ y\\ z \end{smallmatrix}\big) \) describes a corresponding point \(k\) in 3D space and \(N_\text{c}\) is the amount of corresponding points per object, which is object dependent.
We calculate \(d_\text{l}\) in the same manner by treating the reference line as a virtual object.
The shortest point-to-line distance between corresponding object points on the grasped object and the reference line defines the corresponding point on the reference line.

\subsubsection*{\textbf{Adjacency Aware Object Release}}
When the agent decides to open the gripper, it receives the reward \(r_\text{r}\) defined as:
\begin{align}
\label{eq:reward_q1-2}
r_\text{r}
=&
\alpha_\text{c} \cdot
[
\underbrace{(1 - \tanh(\beta \cdot d_\textrm{o}))
+ 2 \cdot (1 - \tanh(\beta \ccdot d_\textrm{l}))}_{\mathclap{\text{placement closeness}}}\nonumber\\
&+ \underbrace{(1 - \tanh(\beta \cdot \Delta\theta_\text{o}))}_{\mathclap{\text{orientation closeness}}}
]
-
\underbrace{\alpha_\text{h} \cdot d_\text{h}}_{\mathclap{\text{drop penalty}}}
\text{.}
\end{align}

The first three summation terms incentivize the agent to optimize the spatial compactness of the assembly, while the last term ensures that the object is not dropped mid-air.
It depends on the height \(d_\text{h}\) during the release.
For the closeness terms, the hyperbolic tangent ensures precise placement through a higher slope in the proximity of neighboring objects and reference lines, while larger distances \(d_{\text{o/l}}\) receive less importance through a smaller slope. 
To ensure a proper alignment, we deem the closeness to reference lines twice as important as that to placed objects.

\subsubsection*{\textbf{Collision Avoidance}}
We penalize the agent for contacts (\(\leftrightarrow\)) with the EE \(EE\), the workspace surface \(w\), the grasped object \(o_\text{g}\), and placed objects \(o_\text{p}\) by the reward \(r_\text{col}\) defined as: 
\begin{align}
\label{eq:reward_collision}
r_\text{col} =
\begin{cases}
- \alpha_\text{o} \text{ if } (o_\text{g} \leftrightarrow o_\text{p}) \vee (EE \leftrightarrow o_\text{p})\\
- \alpha_\text{w} \text{ if } (EE \leftrightarrow w)
\text{.}
\end{cases}
\end{align}

The collision conditions are also termination criteria for the episode.

\section{Dataset Generation}
\label{sec:generator}
For training and evaluation on realistic data, we generate an object dataset, based on the exemplary application of fresco reconstruction.
The dataset consists of plain fresco fragments with irregular shapes as shown in \figref{fig:motivation}.
For the generation, we assume that jigsaw puzzles present a suitable abstraction of archaeological frescoes.
Harel and Ben-Shahar \cite{harelCrossingCutsPolygonal2021} create puzzles by cutting through a global polygonal shape with an arbitrary number of straight cuts.
This method generates entirely convex polygons.
We adapt this idea to generate fresco fragments that are realistic in dimensions, shape, and mass.
Each fresco consists of \(18\)\,-\,\(23\)~fragments with \(11\)\,-\,\(19\)~vertices which enclose an area of \(7.0\)\,-\,\SI{18.1}{cm\squared}.

We generate 3D CAD models to use the generated fragments with common robot simulators such as PyBullet \cite{coumansPyBulletPythonModule2016}
Additionally, the generator computes the assembly sequence by sliding a window with a fixed size in a snake-like pattern over the 2D fresco polygons (see \figref{fig:architecture}\,a).
When a fragment enters the sliding window it becomes part of the sequence.
The method is similar to bottom-left nesting algorithms~\cite{xieNestingTwodimensionalIrregular2007a} used, e.g., in dense packing.

\section{Experimental Evaluation}
\label{sec:exp}
To demonstrate the performance of our RL approach for efficient object placement, including positional and angular displacement relative to the original layout, we conduct a qualitative and quantitative comparison with two baseline approaches.
The supplementary video\footnote{Video: \href{https://youtu.be/_Au5yyJhiP4}{https://youtu.be/\_Au5yyJhiP4}} provides additional details about our approach.
Moreover, our code\footnote{Code: \href{https://github.com/HumanoidsBonn/compact_rl_placement}{https://github.com/HumanoidsBonn/compact\_rl\_placement}} is publicly available on GitHub.

\subsection{Experimental Setup}
\label{subsec:setup}

For all experiments, we use the 6-DoF robotic arm UR5, equipped with the two-finger gripper Robotiq 2F-85, as shown in \figref{fig:motivation}.
However, our approach is applicable to any robotic arm, since our agent learns EE movement increments, and it does not directly learn to control the robot's joint positions.
Furthermore, we rely on the TQC implementation of Stable-Baselines3~\cite{raffinStableBaselines3ReliableReinforcement2021} and the OpenAI Gym toolkit~\cite{brockmanOpenAIGym2016a}, as well as PyBullet~\cite{coumansPyBulletPythonModule2016} for simulation.
The actor and critic networks use three-layered fully-connected networks of size \{\(128, 128, 128\)\}.
We trained our agent using the reward scaling factors in \tabref{tab:factors} and the training parameters in \tabref{tab:training}.
We experimentally determined the scaling factors that lead to the best performance.
To show the agent's generalization capabilities to unseen object shapes, we train it on one fresco and evaluate it on \(100\)~novel frescoes.
The agent was trained for approximately \(4\)\,hours on a desktop computer equipped with an Intel i9-11900KF CPU and \(64\)\,GB RAM, without using a dedicated GPU.

\begin{table}[b]
	\begingroup
	\setlength{\tabcolsep}{2pt}
	\centering
	\caption{Scaling factors used for the evaluation}
	\label{tab:factors}
	\begin{tabularx}{\linewidth}{*{6}{|Y}|}
        \hline
        \(\alpha_\text{n} = 0.1\) & \(\alpha_\text{c} = 2\) & \(\beta = 3\) &  \(\alpha_\text{h} = 0.5\) & \(\alpha_\text{o} = 5\) & \(\alpha_\text{w} = 1\) \\
        \hline
	\end{tabularx}
	\endgroup
\end{table}

\subsection{Ablation and Baselines}
\label{subsec:baselines}
\begin{figure}[t]
	\centering
	\includegraphics[width=0.99\linewidth]{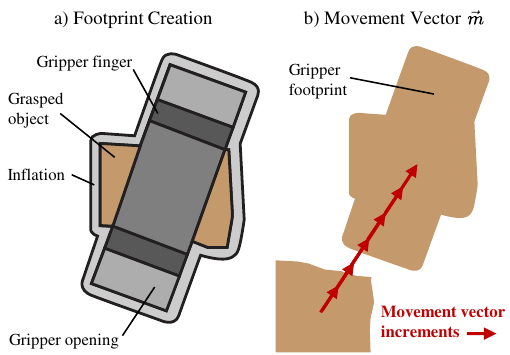}
	\vspace{-1.25\baselineskip}
	\caption{
		a)~We attach the required operational space of the gripper to the grasped object's shape and create their union.
		By inflating the union by \SI{1}{mm}, we add a safety margin for the placement. The resulting shape is the gripper footprint.
		b)~Baseline 2 uses the gripper footprint to find a suitable placement location by incrementally moving the shapes of the placed object and the gripper footprint until they no longer overlap. The movement is defined by the movement vector \(\vec{m}\).
	}
	\label{fig:gripper_footprint}
\end{figure}

In this section, we elaborate on the conducted ablation study of \textbf{OUR} approach and the two baselines approaches we used for comparison.

\subsubsection*{\bfseries{Ablation}} 
\label{subsubsec:ablation}
We perform an ablation of OUR approach by removing the guiding reference lines (\textbf{OUR-ABL}).
Without these lines, the agent lacks information about the assembly's boundaries. 
In the ablation, we remove the reference lines from the observation space and the reward function to understand their effect on the placement performance.

To the best of our knowledge, there is no other work that focuses explicitly on the constrained placement of objects, as explained in \secref{sec:related} and \secref{sec:main}.
Therefore, we compare OUR approach, to two baselines (BL1 and BL2), which are not RL-based, and explained below.

\subsubsection*{\bfseries{Baseline 1}}
\label{subsubsec:baseline1}
For \textbf{BL1}, we generate equal gaps between all objects and check during execution whether the gap size is sufficient to avoid collisions.
For this purpose, we iteratively increase the scale of the complete layout by \(\alpha_\text{b}\)\,\(=\)\,\(0.1\) using an affine transformation and place each unscaled object at the shifted position resulting from the scaling.
Compared to the original layout, the updated version has gaps between all objects.
The gap size per iteration depends on the magnitude of \(\alpha_\text{b}\).
As long as collisions occur during the placement, we repeat the process.
\figref{fig:evaluation}\,c depicts an exemplary placement result of BL1, showing an equally spaced assembly with larger gaps than necessary, due to the unique shapes of the objects.

\subsubsection*{\bfseries{Baseline 2}}
\label{subsubsec:baseline2}
In contrast to BL1, the main idea of \textbf{BL2} is to generate individual gaps based on the objects' shapes between the grasped object and already placed objects, to increase the compactness of the assembly.
Therefore, we add the space required to open the gripper to the shape of the grasped object, as shown from a top-down view in \figref{fig:gripper_footprint}\,a.
We then perform a union of the two shapes and inflate the resultant by \SI{1}{mm}, which adds a safety margin.
\figref{fig:gripper_footprint}\,b shows the resulting shape called the gripper footprint.
The footprint is initially placed next to the previously placed object according to the assembly sequence.
If any shapes overlap in this step, we calculate an incremental movement vector \(\vec{m}_\text{inc}\) to shift the current gripper footprint out of collision.
Given the footprint's centroid \(C_\text{f}\) and all centroids of colliding shapes \(C_{\text{col},i}\), we calculate the total movement vector \(\vec{m}\) as follows: 
\begin{equation}
\label{eq:movement_vector}
\vec{m} =
\sum_{j=0}^{N_\text{inc}} 
\vec{m}_{\text{inc},j}
\text{ with }
\vec{m}_{\text{inc},j} =
\psi \cdot
\sum_{i=1}^{N_{\text{col},j}} 
	\frac{
		C_\text{f} - C_{\text{col},i}
	}{
		|| C_\text{f} - C_{\text{col},i} ||
	}
\text{,}
\end{equation}

\begin{table}[b]
	\centering
	\caption{Notations and Training settings\label{tab:training}}
	\begin{tabularx}{\linewidth}{llX}
		\toprule[\lightrulewidth]
		Notation & Value & Description \\
		\midrule[\lightrulewidth] 
		\(N_\text{ep}\) & 10 & Maximum number of steps per episode\\
		\(N_\text{crit}\) & 2 & Number of critics\\
		\(B_{E}\) & \(1 \cdot 10^{5}\) & Experience replay buffer size\\
		\(B_{G}\) & \num{128} &  Minibatch size for each gradient update\\
		\(l\) & \(1 \cdot 10^{-3}\)  & Learning rate\\
		\(\gamma\) & 0.95 & Discount factor\\
		\(\sigma_{\epsilon_\pi}\) & 0.1 & Std. deviation of exploration noise \(\epsilon_\pi\) \\
		\(\tau\) & 0.05 & Soft update coefficient\\		
		\bottomrule[\lightrulewidth]
	\end{tabularx}
\end{table}

\begin{figure*}[t] 	\centering \includegraphics[width=0.99\linewidth]{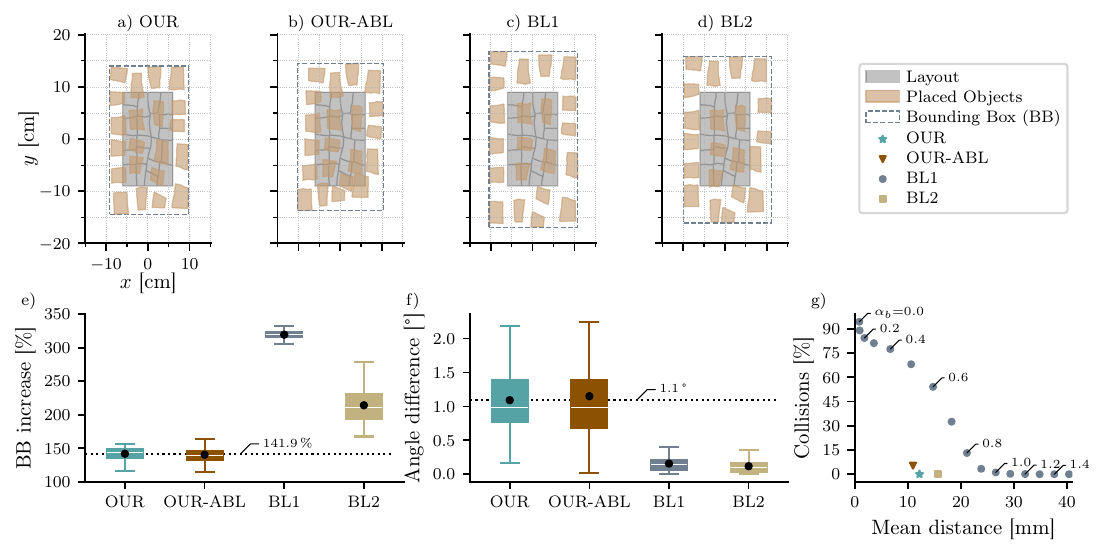}
        \vspace{-0.5\baselineskip}	
        \caption{The plots show all results of the qualitative and quantitative evaluation.
		a-d)~The 2D top-down view shows fresco assemblies abstracted as polygons.
		a)~Using OUR approach results in the most compact assembly.
            b)~Without the reference lines in OUR-ABL, there is a clear object shift and assembly skew, highlighting the necessity of their use.
		c)~Scaling the fresco according to BL1 results in a larger bounding box due to the equal spacing.
		d)~Placing the objects relative to each other according to BL2 results in a higher displacement compared to OUR approach, but lower than BL1.
		%
		%
		e-g)~The plots show the results of the quantitative evaluation.
            e)~OUR approach has the smallest bounding box increase (BBI-I) compared to both baselines.
		f)~The angle difference of both baselines is smaller than OUR approach.
		g)~OUR approach achieves the smallest mean object distance with no collisions.
            The collision rate of BL1 decreases with an increasing scaling factor \(\alpha_\text{b}\).
	} 
	\label{fig:evaluation}
\end{figure*}

where \(j\) is the increment iterator, \(N_\text{inc}\) is the amount of movement increments, \(\psi\) is the increment step size, \(i\) is the collision shape number \((i \in \{1,...,N_{\text{col},j})\), and \(N_{\text{col},j}\) is the amount of shapes the footprint is colliding with.
Note that \(N_{\text{col},j}\) depends on the current increment \(j\), because we move the footprint until no shapes overlap anymore, to determine the suitable placement location, as shown in \figref{fig:gripper_footprint}\,b.
Also, note that \figref{fig:gripper_footprint}\,b depicts solely one collision shape \(C_{\text{col},1}\).
\figref{fig:evaluation}\,d shows an exemplary placement result of BL2.

\begin{table}[t]
    \begingroup
	\setlength{\tabcolsep}{2pt}
	\centering
	\caption{Results of the quantitative evaluation}
	\label{tab:evaluation}
    	\begin{tabularx}{\linewidth}{l*{4}{|Y}}
    		\toprule[\lightrulewidth]
    		\multicolumn{1}{c}{Metric} & \multicolumn{1}{c}{OUR} & \multicolumn{1}{c}{OUR-ABL} & \multicolumn{1}{c}{BL1} & \multicolumn{1}{c}{BL2} \\
    		\midrule[\lightrulewidth]
    		a)\,BBI-I\,[\%]\,\textdownarrow & \scriptsize \(141.9 \pm 9.4\) & \scriptsize \(140.4 \pm 11.8\) & \scriptsize \(319.4 \pm 6.6\) & \scriptsize \(214.2 \pm 26.7\) \\
    		b)\,BBI-FW\,[\%]\,\textdownarrow & \scriptsize \(26.0 \pm 4.6\) & \scriptsize \(25.3 \pm 6.0\) & \scriptsize \(118.5 \pm 3.5\) & \scriptsize \(63.6 \pm 13.2\) \\
    		c)\,Skew\,[°]\,\textdownarrow & \scriptsize \(2.0 \pm 1.1\)& \scriptsize \(3.0 \pm 0.8\) & \scriptsize \(1.0 \pm 0.7\) & \scriptsize \(3.9 \pm 3.1\) \\
    		d)\,Angle\,[°]\,\textdownarrow & \scriptsize \(1.1 \pm 0.5\) & \scriptsize \(1.1 \pm 0.8\) & \scriptsize \(0.2 \pm 0.1\) & \scriptsize \(0.1 \pm 0.1\) \\
    		e)\,Dist.\,[mm]\,\textdownarrow & \scriptsize \(12 \pm 1\) & \scriptsize \(11 \pm 1\) & \scriptsize \(35 \pm 1\) & \scriptsize \(16 \pm 2\) \\
    		f)\,Coll.\,[\%]\,\textdownarrow & \scriptsize \(0.0 \pm 0.0\) & \scriptsize \(5.3 \pm 5.7\) & \scriptsize \(0.0 \pm 0.0\) & \scriptsize \(0.0 \pm 0.0\) \\
    		\bottomrule[\lightrulewidth]
            \end{tabularx}
    \endgroup
    \vspace{1ex}
    
    \noindent
    \textbf{Metrics legend:} 100 novel assemblies are the basis for the metrics. \textdownarrow\,\,Lower values are preferable to higher values. The ideal bounding box increase (BBI-I) measures the difference between the minimum rotated rectangle area encompassing all placed assembly objects and the ideal layout area divided by the ideal layout area. For BBI-FW, we scale the ideal layout so that the width of the gripper finger fits between all object shapes. The skew (Skew) assesses the mean skew angle between the \(x\) and \(y\) bounding box axes and the placed assembly's \(x\) and \(y\) axes. The mean circular angle difference (Angle) measures the difference between the object angles defined in the layout and all placed objects. The mean object distance (Dist.) calculates the minimum centroid distances between all placed objects. The collision rate per assembly (Coll.) checks the amount of contacts between all placed objects and between the robot and the placed objects.
\end{table}

To receive comparable results among all approaches and the ablation, we assign a randomly generated grasp yaw angle \(\theta_{\text{grasp}} \in [-\SI{90}{\degree},+\SI{90}{\degree}]\) to each object.
During all experiments, \(\theta_{\text{grasp}}\) determines the object specific grasp.
For action execution of each baseline, we use an inverse kinematics pipeline to generate collision-free motions to the placement targets computed by the baselines.
Since the performance of BL1 depends on the scaling factor \(\alpha_\text{b}\), we experimentally determined the smallest scaling factor (\(\alpha_\text{b}\)\,=\,\(1.3\)) where no collisions appeared during the execution of all assemblies.
For BL2, we experimentally determined an increment step size of \(\psi\)\,=\,\SI{3}{mm}.

\subsection{Experimental Results}
\label{subsec:sim_results_qual}

We provide both a qualitative and a quantitative comparison of OUR method with OUR-ABL ablated method and with the two baselines (BL1 and BL2).
\figref{fig:evaluation} and \tabref{tab:evaluation} summarize the experimental results.
Based on averages of 100 assemblies, we calculate the six different metrics detailed below to measure the assembly efficiency.

\subsubsection*{\textbf{Metrics}}
We first calculate the minimum rotated rectangle area encompassing all placed objects, referred to as BB.
The bounding box increase (\textbf{BBI-I}) measures the difference between BB and the ideal layout area divided by the ideal layout area.
For \textbf{BBI-FW}, we inflate each individual object shape in the ideal layout by the width of the gripper finger and incrementally move the inflated shapes out of collision by scaling the assembly like in BL1. 
Then, we calculate the bounding box area of the resulting layout.
BBI-FW measures the difference between the BB and the finger width inflated layout, in a manner similar to BBI-I.
Note that the BBI-FW reference layout is a geometric metric and does not take real-world constraints like object placement imprecisions into account.
The skew (\textbf{Skew}) assesses the mean absolute angle difference between two skew lines of the placed assembly and the \(x\) and \(y\) axes of the bounding box.
We determine the skew lines using three extreme points in the convex hull of the placed assembly.
The left skew line connects the bottom-left point with the upper-left point and the bottom skew line connects the bottom-left point with the bottom-right point.
The mean circular angle difference (\textbf{Angle}) measures the difference between the object angles defined in the layout and all placed objects.
The mean object distance (\textbf{Dist.}) computes the minimum centroid distances between all placed objects.
The collision rate (\textbf{Coll.}) per assembly evaluates the number of contacts between all objects and between the robot and the placed objects.
For the different metrics, we performed the two-sided Mann-Whitney U test~\cite{mannTestWhetherOne1947} to check for statistical significance between the approaches. 

\subsubsection*{\textbf{Ablation Results}}
We observed that the reference lines enable OUR method to demonstrate a significant (\({p < 0.001}\)) reduction in the skewness with respect to the bounding boxes compared to OUR-ABL.
As can been seen in \figref{fig:evaluation}\,a-b, the qualitative plots show that removing the reference lines leads to a more unstructured assembly compared to OUR approach with uneven gaps between the pieces while also shifting the pieces in both the horizontal and the vertical direction. 
The unstructured assembly also leads to \SI{5.3}{\percent} of the pieces per assembly experiencing collisions in OUR-ABL method, whereas no collisions occur in OUR method. 
While OUR-ABL performs similarly on all other metrics compared to OUR method, the occurrence of collisions during the assembly makes it unusable for compact placement of objects close to each other. 

These results highlight the necessity of reference lines in OUR method to maintain a balance between compactness and collision-free assembly, and it illustrates their integral role for a successful object placement.

\subsubsection*{\textbf{Baseline Results}}
As can been seen in \figref{fig:evaluation}\,c, BL1 scales the layout uniformly, resulting in a viable, but highly inefficient spatial configuration. 
The bounding box metrics validate this as well, where BL1 has the highest BBI-I of \SI{319.4}{\percent} and BBI-FW of \SI{118.5}{\percent} on average. 
With a mean BBI-I of \SI{214.2}{\percent} and BBI-FW of \SI{63.6}{\percent}, BL2 results in a more compact assembly (see \figref{fig:evaluation}\,d) as compared to BL1. 
However, it is still highly inefficient consuming more than thrice the space needed for the ideal layout. 
The increase in bounding box area is due to the increase in gaps between the pieces which is as high as \SI{35}{mm} for BL1, and \SI{16}{mm} for BL2. 

OUR approach significantly (\({p < 0.001}\)) reduces the bounding box area (see \tabref{tab:evaluation}\,a-b), with a BBI-I of \(\SI{141.9}{\percent} \pm 9.4\), resulting in a more compact placement of the layout.
It is especially important to note that OUR approach has a BBI-FW of only \SI{26.0}{\percent}.
This shows OUR approach is close to the most efficient feasible configuration without any gripper and object collisions. 
In particular, OUR approach improves the space utilization per BBI-I on average by \SI{177.5}{\percent} compared to BL1 and by \SI{72.3}{\percent} compared to BL2. 
This reduction in bounding box area is due to the statistically significantly (\({p < 0.001}\)) lower distance between the objects which is only \SI{12}{mm} on an average for OUR method.

It can been seen that both OUR and OUR-ABL method learn to rotate the objects such that the agent achieves the desired orientation within an accuracy of \SI{1.1}{\degree}. 
Although the baselines BL1 and BL2 have a better orientation accuracy of \SI{0.2}{\degree} and \SI{0.1}{\degree} respectively, the real-world differences would be negligible due to sensor and actor noise. 
On the contrary, OUR method learns to orient the object intelligently depending on the adjacent objects which significantly reduces the average distance between the pieces while minimally sacrificing orientation accuracy. 

Both the baselines and OUR method do not suffer from any collisions, while OUR-ABL method does show the occurrence of collisions, as mentioned in the ablation results.

With regards to the skewing of the object placement along the axes, it can been seen that BL1, being a naive isotropic expansion approach, suffers the least skew of \SI{1.0}{\degree}.
OUR-ABL and BL2 demonstrate a high degree of skew of \SI{3.0}{\degree} and \SI{3.9}{\degree} respectively, whereas OUR method shows the least skew after BL1 with \SI{2.0}{\degree}. 

The quantitative results demonstrate that OUR method is significantly better at ensuring a compact multi-object placement as compared to both the baselines, while suffering no collisions like OUR-ABL, and demonstrating only a slight degradation in orientation accuracy and skewness. 

\section{Conclusion}
\label{sec:conclusion}
In this work, we presented a novel reinforcement learning-based method for the compact placement of irregularly shaped objects that maintains spatial constraints with adjacent objects in the assembly.
Our approach ensures significantly minimized layout inflation during placement, measured by the bounding box increase, compared to two baseline methods. 
Importantly, our method achieves compactness close to the geometrically possible layout considering gripper finger width constraints, while achieving zero collisions. 
Furthermore, it maintains dense object assemblies, achieving a notable reduction in mean distances between objects.

%
\bibliographystyle{IEEEtran}
\bibliography{bibliography}
\balance
\end{document}